\documentclass[letterpaper, 10 pt, conference]{ieeeconf}  

\IEEEoverridecommandlockouts                              

\usepackage{graphics} 
\usepackage{epsfig} 
\usepackage{mathptmx} 
\usepackage{times} 
\usepackage{amsmath} 
\usepackage{amssymb}  
\usepackage{booktabs}
\usepackage{multirow}
\usepackage{algpseudocode}
\usepackage{float}
\usepackage{subfigure}
\usepackage[linesnumbered,ruled]{algorithm2e}
\usepackage{graphicx}
\usepackage{textcomp}
\usepackage{xcolor}
\usepackage{stackengine}
\DeclareMathOperator*{\argmax}{arg\,max}

\title{\LARGE \bf
Learning to Generate Unambiguous Spatial Referring Expressions for Real-World Environments
}

\author{Fethiye Irmak Do\u{g}an$^{1}$, Sinan Kalkan$^{2}$ and Iolanda Leite$^{1}$
\thanks{$^{1}$Fethiye Irmak Do\u{g}an and Iolanda Leite are with the Division of Robotics, Perception and Learning from the School of Electrical Engineering and Computer Science at KTH Royal Institute of Technology, Stockholm, Sweden
        {\tt\small \{fidogan, iolanda\}@kth.se}}%
\thanks{$^{2}$Sinan Kalkan is with the KOVAN Research Lab at the Department of Computer Engineering, Middle East Technical University, Ankara, Turkey
        {\tt\small  skalkan@metu.edu.tr}}%
}

\begin{document}

\maketitle
\thispagestyle{empty}
\pagestyle{empty}

\begin{abstract}

Referring to objects in a natural and unambiguous manner is crucial for effective human-robot interaction. Previous research on learning-based referring expressions has focused primarily on comprehension tasks, while generating referring expressions is still mostly limited to rule-based methods. In this work, we propose a two-stage approach that relies on deep learning for estimating spatial relations to describe an object naturally and unambiguously with a referring expression. We compare our method to the state of the art algorithm in ambiguous environments (e.g., environments that include very similar objects with similar relationships). We show that our method generates referring expressions that people find to be more accurate ($\sim$30\% better) and would prefer to use ($\sim$32\% more often). 

\end{abstract}
\maketitle
\section{Introduction}

Verbal communication is a key challenge in human-robot interaction. 
Humans are used to reasoning and communicating with the help of referring expressions, defined as \textit{``any expression used in an utterance to refer to something or someone (or a clearly delimited collection of things or people), i.e., used with a particular referent in mind.''} \cite{hurford2007semantics}. Referring expressions are commonly used to describe an object in terms of its distinguishing features and spatiotemporal relationships to other objects.

The ability to generate spatial referring expressions is critical to many robotics applications, such as to clarify an ambiguous user request to pick up an object or to generate verbal instructions. When there are multiple objects that may fit a single description or multiple ways to describe the same target object (such as in Figure \ref{fig:example}), it is important that the referring expression used by the robot is not only accurate but also similar to what a human would use to facilitate communication and achieve comprehension.

\begin{figure}
\centering
\includegraphics[width=0.49\textwidth]{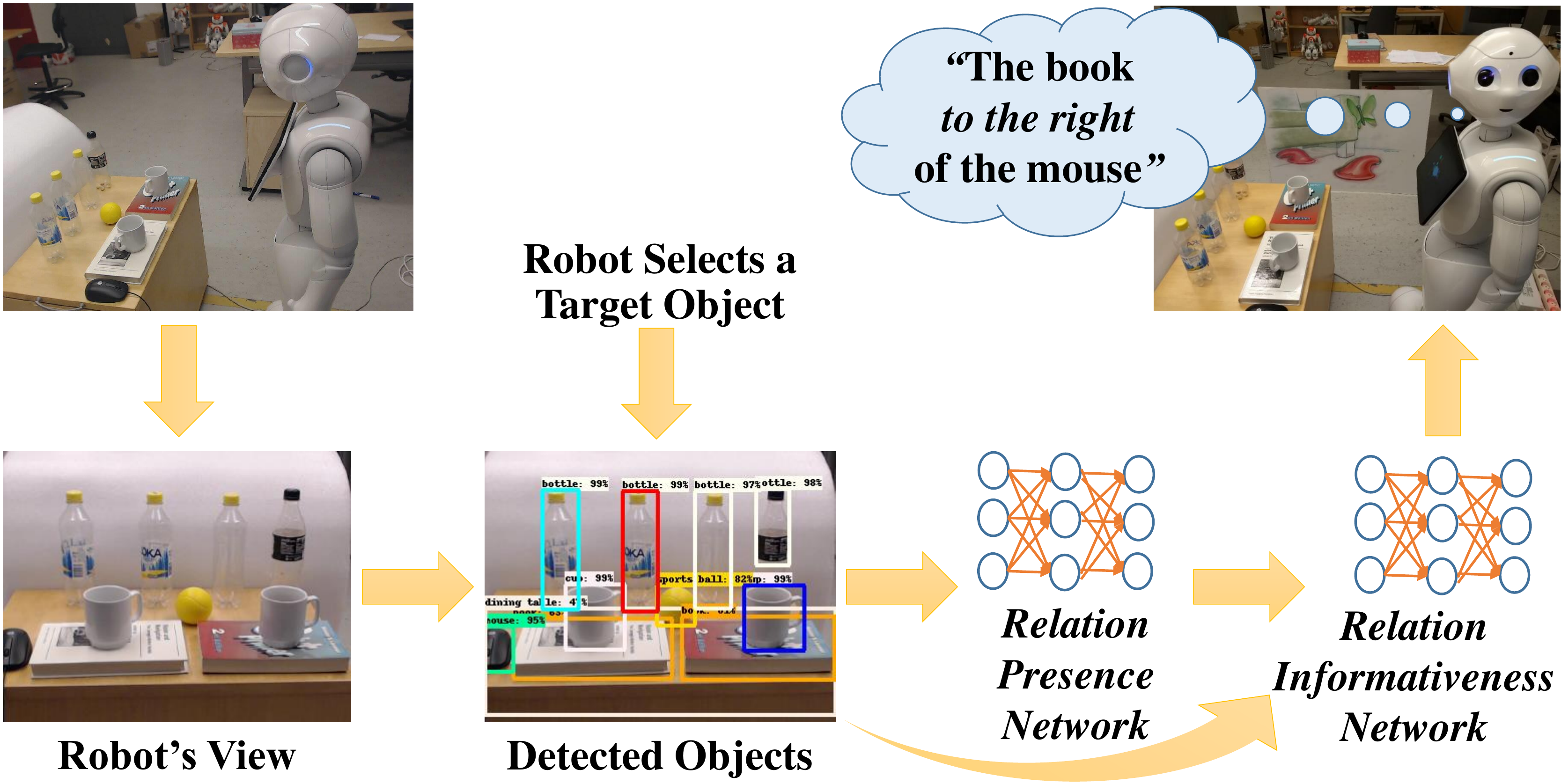}
\caption{Overview of our referring expression generation method.}
\label{fig:overview}
\end{figure}

In this paper, we address the problem of generating unambiguous and natural-sounding spatial referring expressions by using a learning-based method that can be used by robots to describe objects in real-world environments (see Figure \ref{fig:overview}). 
There have been several efforts for both the comprehension and generation of referring expressions for Human-Robot Interaction (HRI). For comprehension, recent works have shown promising results by using the advances in deep learning \cite{hatori2018interactively,shridhar2017grounding,shridhar2018interactive}. Generating spatial referring expressions can be more challenging than comprehension because there can be multiple ways to refer to a target object in relation to other objects, 
yet some descriptions might appear unnatural to human users or not describe the target object in a unique manner. Previous research in generating referring expressions has mostly focused on hand-designed thresholds, rules or templates \cite{williams2017referring,williamsreferring,kunze2017spatial,zender2009situated}, which makes these approaches difficult to generalize to other environments. Computer vision researchers have been addressing this problem using learning-based methods \cite{mao2016generation,yu2017joint,cirik2018using}.

Previous research on spatial referring expression generation has either been less focused on the naturalness of the generated expressions or has not been tested in highly ambiguous scenes. To the best of our knowledge, our method is the first in doing so, while bringing the following benefits over pure rule-based approaches: (i) being able to determine the relations between objects, (ii) learning the confidence of each spatial relation to generate the most natural referring expression for the target object. As we demonstrated with a user study in challenging indoor and outdoor scenes, our method yields more accurate referring expressions that humans would be more willing to use compared to an algorithm by Kunze et al. \cite{kunze2017spatial}.

\subsection{Background}

Referring expressions have been addressed in various HRI and computer vision studies, both in terms of comprehension and generation. Many of these approaches have leveraged spatial relationships. 
 

For the comprehension of referring expressions in HRI, many studies have tried interpreting natural language descriptions from humans for finding target objects or locations. These studies primarily used natural language parsing methods or graph-based representation and search algorithms \cite{zender2009situated,kruijff2007incremental,kollar2013learning,paul2016efficient}. More recently, some works have employed deep learning methods:
For example, Hatori et al. \cite{hatori2018interactively} proposed an interactive system where a deep network processes unconstrained spoken language and maps the words to actions and objects in the scene. Similarly, Shridhar \& Hu \cite{shridhar2017grounding, shridhar2018interactive} used recurrent neural networks to link visual features with provided referring expressions (e.g., `the red can next to the teddy bear') to detect the referred object and to ask disambiguation questions. Both of these studies address comprehension of referring expressions as a learning problem, and reduce the limitations of hand-designed features and rules.



Generating referring expressions is a more complex issue. Because the comprehension tasks have a limited number of possible solutions (i.e., they are bounded by the number of objects present), generation problems are more difficult to solve than comprehension.  Consequently, generating referring expressions has mostly been resolved through hand-designed thresholds, rules, or templates \cite{williams2017referring,williamsreferring,kunze2017spatial,zender2009situated}.
For instance, Williams \& Scheutz \cite{williams2017referring,williamsreferring} proposed a method that extends an incremental algorithm \cite{dale1995computational} to generate domain-independent referring expressions under uncertainty, but they employ rules for generating expressions and pre-determined thresholds for handling uncertainties. In another study, Kunze et al. \cite{kunze2017spatial} used a method with five different algorithm strategies for generating spatial referring expressions and trained a classifier to determine the most appropriate one for a specific scene. Although the classifier learns to decide the algorithm strategy, each algorithm strategy depends on different rules for generating referring expressions. The main disadvantages of rule-based methods include: (i) the assumption of ``perfect, complete, and accessible knowledge of all referents'', which is not always possible 
\cite{williams2017referring}; (ii) the impracticality of hand-designing thresholds and rules that are supposed to generalize to every possible setting; and (iii) that methods are not adaptable/extensible for a life-long learning robot.

\begin{figure}
\centering
\includegraphics[width=0.28\textwidth]{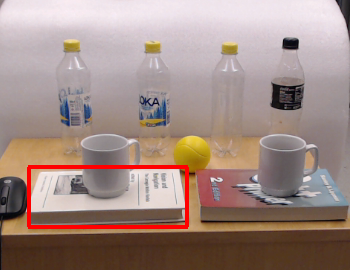}\\ \footnotesize
\textbf{W.r.t. the closest object:}  `The book at the bottom of the cup' \newline \hspace{\textwidth}
\textbf{KRREG \cite{kunze2017spatial}:} `The book to the left of the sports ball' \newline \hspace{\textwidth}
\textbf{Our method:} `The book to the right of the mouse' \newline \hspace{\textwidth}
\caption{An illustration of how challenging generating a referring expression can be. In this example, there are two books in the scene and one can refer to the book in red in many unnatural and ambiguous ways. We focus on selecting unambiguous and natural spatial relations for referring to a selected object.\label{fig:example}}
\end{figure}

There are promising studies on generating unambiguous referring expressions in computer vision \cite{mao2016generation,yu2017joint,cirik2018using}. However, 
these models have not been tested in highly ambiguous scenes (see Figure \ref{fig:example} for an example), and they do not focus on generating human-like referring expressions. Therefore, these models are inappropriate for real-time HRI. 


Referring expressions commonly exploit spatial relations \cite{shridhar2017grounding,kunze2017spatial,viethen2008use}, and this approach has been employed by different HRI studies \cite{kunze2017spatial,moratz2006spatial, guadarrama2013grounding, skubic2004spatial}. However, studies that leverage spatial relations are generally based on rule-based approaches, limited numbers of relationships, or artificial data \cite{viethen2008use,zhang2009rule, viethen2013graphs}. Unfortunately, rule-based approaches  are not expandable for different relations and some relations are difficult to formalize in terms of rules, especially while using 2D data (Figure \ref{boxes}). Moreover, artificial data is not always suitable because, in real environments, the relations might satisfy different rules but one might be the dominant choice for humans. 
Although there are inspiring vision studies that learn spatial relations among objects \cite{malinowski2014pooling,haldekar2017identifying}, they either assume prior knowledge about the target and reference objects or they do not have any learning on spatial relation which describes the object unambiguously and naturally.


\subsection{Contributions}
We summarize our contributions in this paper as follows:
\begin{itemize}
\item Rather than hand-designing rules for every relation, our method is capable of learning relations between objects and deducing the dominant one when multiple relationships exist. 
\item Our method learns the informativeness of each spatial relation, i.e., the value of a relation in describing an object with respect to another without ambiguity, and generates the most natural and unambiguous referring expression to describe the target object.
\item Our work is applicable to different indoor and outdoor environments and employable for different HRI tasks. 


\end{itemize}

\section{Referring Expression Generation}
\label{sect:REG}

\begin{figure}
\centerline{
\includegraphics[width=0.35\textwidth]{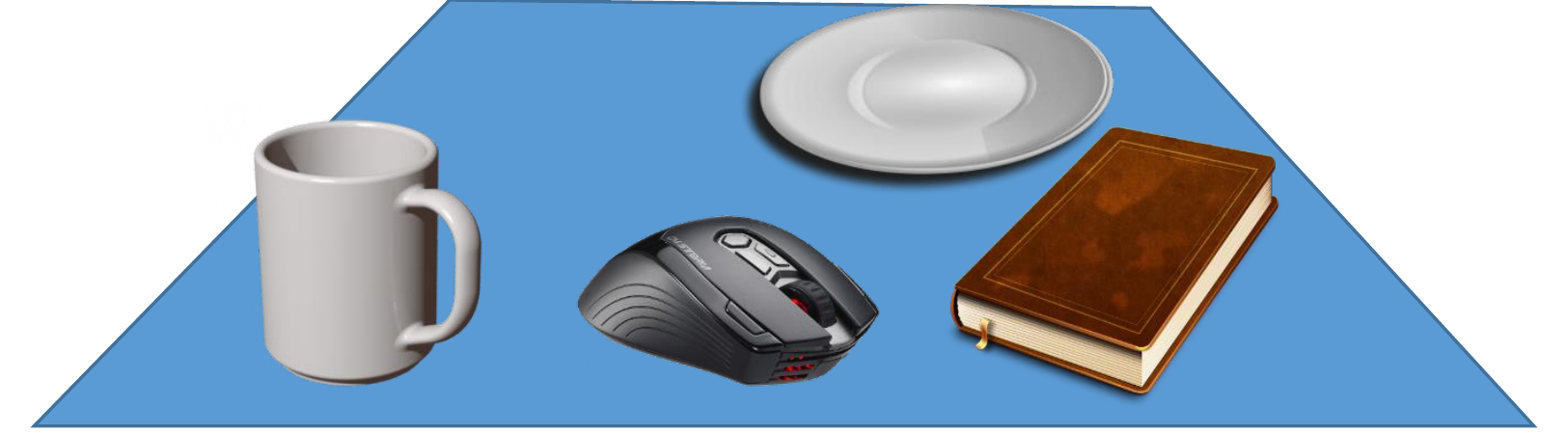}
}
\caption{An example illustrating the motivation behind having one network for the presence of a relation and another network to assess its informativeness. Above, both the mouse and the book are to the right of the cup, but this relationship does not exist between the cup and the plate. The first network detects this presence. Next, it is more informative to use the mouse for referring to the cup in this example because it is the closest. The second network detects this informativeness. \label{fig:two_stage_motivation}}
\end{figure}
We define the referring expression generation (REG) problem to be generating a noun phrase describing a target object $o_t$ with respect to a reference object $o_r$ in terms of their spatial relations (see Figure \ref{fig:int} for examples). With this problem formulation, we are limiting our approach to referring expressions involving two objects and descriptions involving spatial relations only.

We considered the following spatial relations: $\mathcal{R}=\{$`to the right', `to the left', `on top', `at the bottom', `in front', `behind'$\}$, which can easily expand if there is labeled data. By using one of these relations in $\mathcal{R}$, we aim to generate an unambiguous and natural referring expression for a target object, $o_t$, in an encountered scene.

Given an image of a scene, we first detect and find the set of objects, $\mathcal{O}$, using a deep network. If there is a spatial relation of category $r\in\mathcal{R}$ between two objects $o_i$ and $o_j$, we denote it by $s_{ij} =r$. Given the bounding boxes and the types of the objects, we use two networks to find the most informative spatial relation, $s$, and the object to describe a target object, $o_t$ (see Figure \ref{fig:overview}):
\begin{itemize}
\item \textbf{Relation Presence Network (RPN):} The network is trained on a pair of objects using their bounding box coordinates and decides which relations exist between the pair. That is, the RPN network gives us a probability, $p(s_{ij}) \in [0,1]$, for the presence of spatial relation between objects $o_i$ and $o_j$.

\item \textbf{Relation Informativeness Network (RIN):} This network is trained to decide whether a given relation is informative for a pair of objects. This network yields an informativeness measure, $c(s_{ij}) \in [0,1]$, for the given pair of objects $o_i$ and $o_j$ and the spatial relation $s_{ij}$.
\end{itemize}

The motivation behind our two-stage approach is illustrated in Figure \ref{fig:two_stage_motivation}. This approach for finding an informative relation and an object for referring to a target object is similar to the two-stage object detection approaches (e.g., \cite{huang2017speed}), where one network (stage) is devoted to proposing image regions that are likely to include an object and another is employed to classify those regions into object types.



\subsection{Detecting Objects in the Scene}

\begin{figure}
\centering
\includegraphics[width=0.32\textwidth]{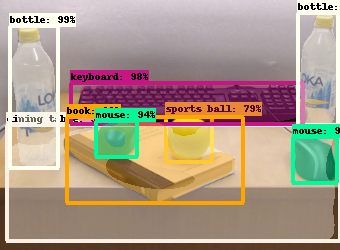}
\caption{An example scene with objects detected by Mask R-CNN \cite{huang2017speed}.}
\label{fig:bounding}
\end{figure}

To detect the objects in a scene, we used Mask R-CNN, one of the state-of-the-art object detectors based on convolutional neural networks \cite{huang2017speed}. Our Mask R-CNN model uses the ResNet-101 as the backbone network and is trained on the MS COCO object detection dataset. This network extracts bounding boxes for each object as well as their types (e.g., book, mouse, etc.) -- see Figure \ref{fig:bounding} for an example.

We describe an object $o_i$ as a quadruple of $<x, y, w, h>$, where $(x,y)$ is the position of the top-left corner of the bounding box containing the object and $w, h$ are the box's width and height.

\subsection{Relation Presence Network (RPN)}

\begin{figure}
\centering
\includegraphics[width=0.47\textwidth]{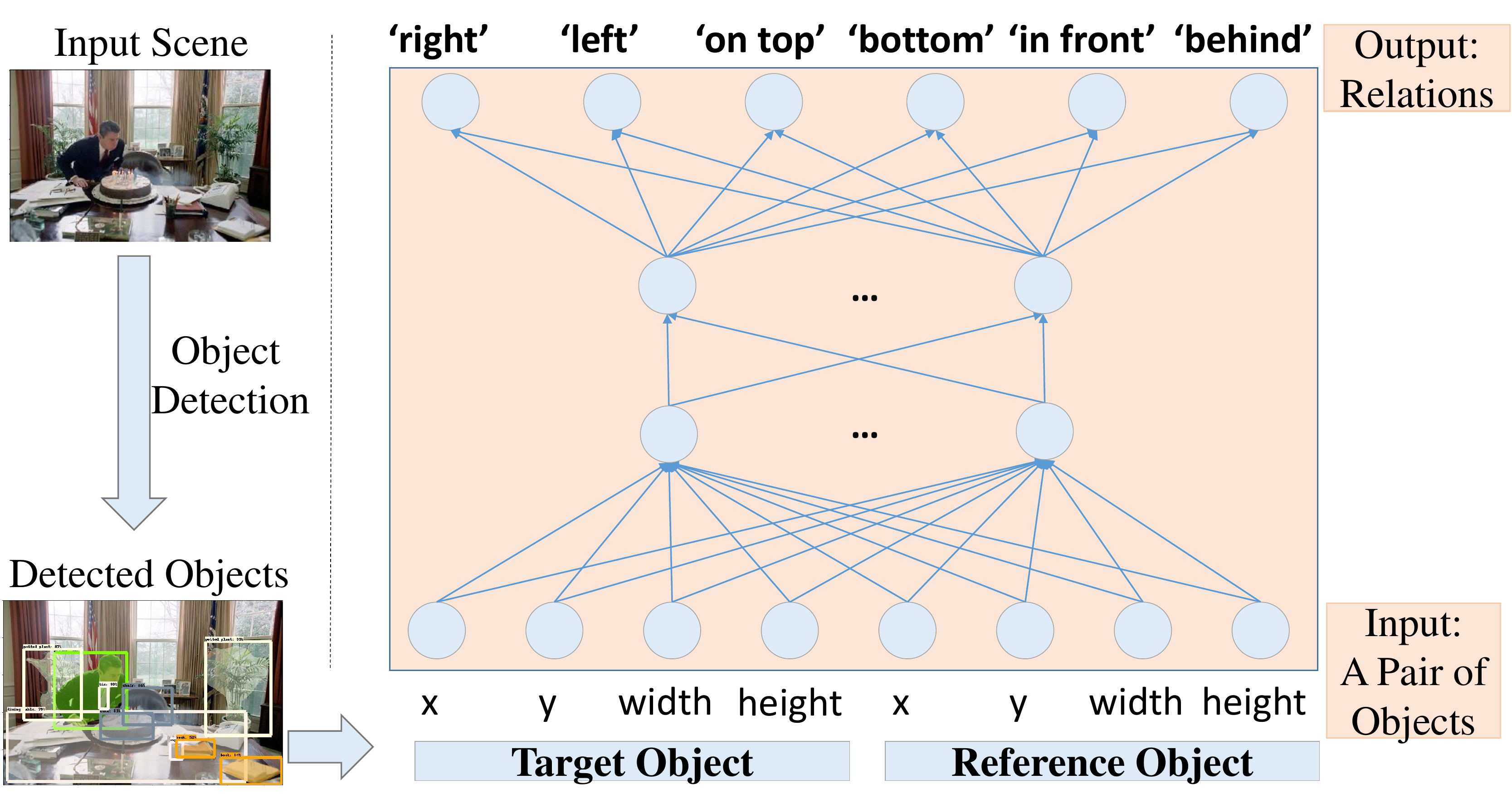}
\caption{Structure of the Relation Presence Network (RPN).}
\label{fig:REN}
\end{figure}

We formulated relation presence estimation as a multi-class classification problem. To this end, we trained a multilayer perceptron, as shown in Figure \ref{fig:REN}. The input of the network is a pair of objects: the target object, $o_t$, and a reference object, $o_r$. The input is provided as a concatenation of the sizes and locations of the bounding boxes of these two objects in an 8-dimensional input space. 
The network has six outputs that contain the probabilities of each possible spatial relation (i.e., $p(s_{ij})$).

RPN has two hidden layers with 32 and 16 hidden neurons, respectively, with ReLU non-linearities \cite{nair2010rectified}. The output layer uses a softmax function to map activations to probabilities. Multi-label cross-entropy loss is used to train the network. Moreover, dropout \cite{srivastava2014dropout} is added between each layer including input and output layers with 0.2 probability to prevent overfitting. Early stopping is used to stop training whenever validation accuracy stop increasing.

The network is trained on a subset of the Visual Genome dataset \cite{krishnavisualgenome}, which includes a rich set of objects and annotations of the spatial relations among them. 

\subsection{Relation Informativeness Network (RIN)}

We formulated informativeness estimation as a binary classification problem, that is, a problem that asks whether or not a spatial relation ${s_{ij}}$ is informative. The confidence of the answer is taken as the informativeness $c_{s_{ij}}$ of the spatial relation. For this purpose, we trained another multilayer perceptron. The input for this network is 12 dimensional: the target object, $o_t$, the reference object, $o_r$ and the relation category (one-hot vector). The network has a single output for the informativeness of the relation between these objects. 

RIN has three hidden layers with 64, 16, and 8 hidden neurons. The hidden and output layers have ReLU and sigmoid nonlinearities, respectively. In RIN, dropout with 0.2 probability is added to each layer. Early stopping is used to stop training when validation accuracy stops increasing.


RIN was trained on another subset of the Visual Genome dataset. A relation is considered as `informative', i.e., more natural, when the relation is annotated in the dataset. Furthermore, it is regarded as `non-informative', i.e., less natural, when that relation exists but it is not annotated. The dataset used for the training of RIN is detailed in Section \ref{data}. 


\subsection{Forming a Referring Expression for a Pair of Objects}

Given a target object ($o_t$) to refer to, our goal is to find the noun phrase $S$ describing $o_t$ in relation to a reference object, $o_r$, with a spatial relation $s_{tr}$ such that $S$ is as informative as possible.

\begin{figure}
\centerline{
\subfigure[`The cup behind the sports ball']{
	\includegraphics[width=0.22\textwidth]{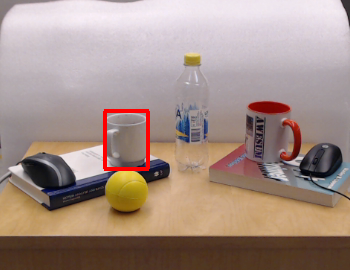}
	}
    \subfigure[`The mouse on top of the book']{
	\includegraphics[width=0.22\textwidth]{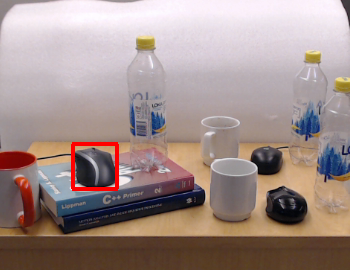}
	}
}
\centerline{
 \subfigure[`The chair to the right of the couch']{
	\includegraphics[width=0.22\textwidth]{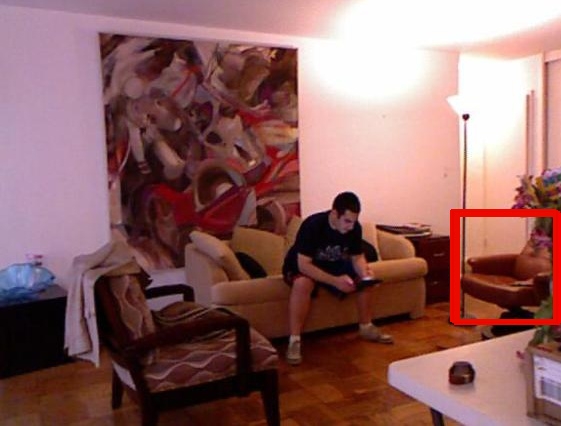}
	}
 \subfigure[`The person on top of the car']{ 
	\includegraphics[width=0.22\textwidth,height=0.135\textheight]{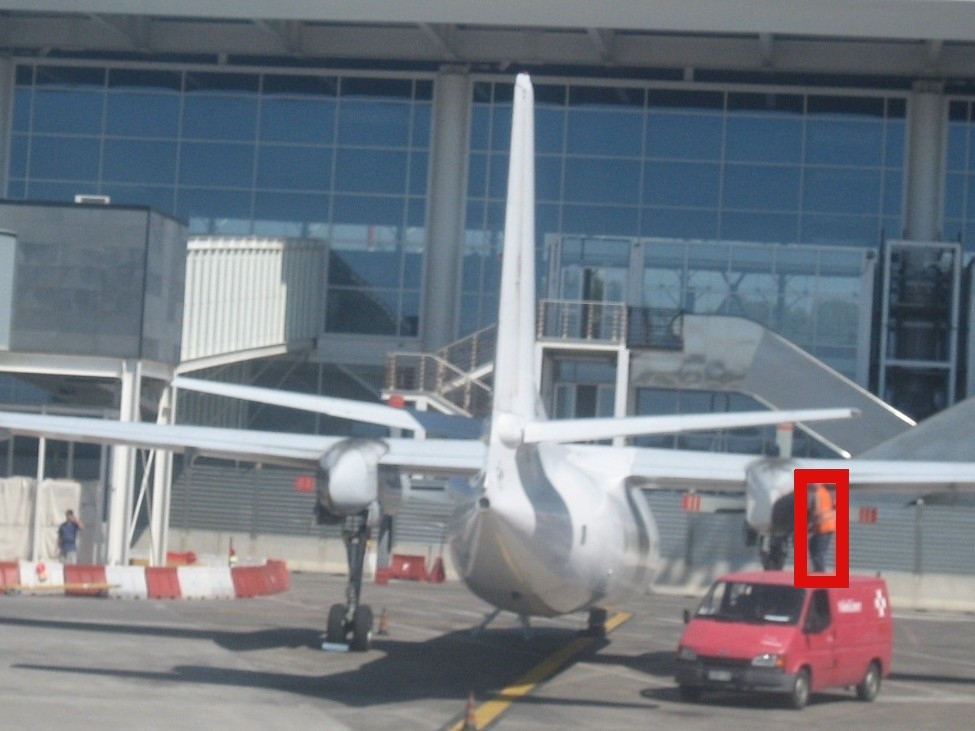}
	}
}
\caption{Referring expressions generated by our method from different indoor and outdoor scenes.}
\label{fig:int}
\end{figure}

%
%
%
First, for each object $o_i$, we select the set of relations, denoted by $\mathcal{S}_i$,  whose presence indicators (provided by the RPN network) are strong (i.e., above the threshold):
\begin{equation}
\mathcal{S}_{i} = \{s_{ij}\ |\ p(s_{ij}) > T\ \textrm{for $s_{ij}$ between $o_i$ and $o_j$}\},
\label{eqn:rel}
\end{equation}
{\noindent}where the threshold $T$ is set to $0.5$ empirically. Even in simple scenes, multiple relations turn out to be confident for each category (for an example, see Figure \ref{boxes3}). Among these, we select the relations with the highest confidence values provided by the RIN network for each category as follows: 
\begin{equation}
\mathcal{S}^*_{i} = \left\{\argmax_{s_{ij}\in \mathcal{S}_i\ \wedge\ s_{ij}=r} c(s_{ij}), \quad \forall r \in \mathcal{R} \right\}.
\label{eqn:S*}
\end{equation}

We select a subset of  $\mathcal{S}^*_{i} $, denoted by  $\mathcal{S}^*_{u}$, such that  $\mathcal{S}^*_{u}$ contains all relations in $\mathcal{S}^*_{i} $ except the relations in $\mathcal{S}_t$:
\begin{equation}
\mathcal{S}^*_{u} = \{s_{ij}\ | s_{ij} \in \mathcal{S}^*_{i}\ and\ s_{ij} \notin  \mathcal{S}_{t} \},
\label{eqn:Su*}
\end{equation}
{\noindent}where $\mathcal{S}_{t}$ is the set of $s_{ti}$ where relation $s_{ti}$ holds between objects $o_t$ and $o_i$. 

For the target object $o_t$, we check whether a relation $s_{ti}$ with object $o_i$ resembles another confident relation in $\mathcal{S}^*_{u}$. In other words, we look if there is a spatial relation $s_{uv}\in \mathcal{S}^*_{u}$ such that $n(o_t)=n(o_u)$ and $n(o_i)=n(o_v)$, where $n(\cdot)$ is the type of the object.  If there is such a relation, we remove $s_{ti}$ from $\mathcal{S}_t$, the candidate relations because its description of $o_t$ is ambiguous.



After eliminating ambiguous relations, we pick the most confident remaining relation to refer to $o_t$ as follows:
\begin{equation}
s_{tr} \leftarrow \argmax_{s_{ti} \in \mathcal{S}_t} c(s_{ti}).
\label{eqn:str}
\end{equation}

Given the target object $o_t$, the reference object $o_r$ and the most informative spatial relation $s_{tr}$ between them, we form a simple noun-phrase by concatenating names of their types $n(\cdot)$ -- see Figure  \ref{fig:int} for some examples.

The overall procedure is summarized in Algorithm \ref{alg:all_model}.

\begin{algorithm}[hbt!]
	\caption{Learning-based method for generating a referring expression.
		\label{alg:all_model}}
        \KwIn{$o_t$, the target object to be described: A vector containing the upper left corner position, the width, and the height of the target object bounding box (obtained from Mask R-CNN).}
         \KwOut{
         	$S$, the referring expression.
        }
		Detect all objects in the scene using Mask R-CNN\\
        For each possible relation $s_{ij}$ between object $o_i$ and $o_j$, determine $p(s_{ij})$ and the confidence $c(s_{ij})$ using the RPN and RIN networks\\
        Set $\mathcal{S}_{i}$ to be the set of $s_{ij}$, such that relation $s_{ij}$ holds between object $o_i$ and $o_j$ (Eq. \ref{eqn:rel})\\
        Let $\mathcal{S}_{t}$ to be the subset of $\mathcal{S}_{i}$, the set of relations between object $o_t$ and $o_i$\\
        Find $\mathcal{S}^*_{i}$, the most confident relations for each relation category for object $o_i$ (Eq. \ref{eqn:S*})\\
        Select $\mathcal{S}^*_{u}$, all relations in $\mathcal{S}^*_{i}$ except the relations in $\mathcal{S}_t$ (Eq. \ref{eqn:Su*})\\
        \If{$n(o_t) = n(o_u)$ and $n(o_i) = n(o_v)$, for relations $s_{ti}\in \mathcal{S}_t$ and $s_{uv} \in \mathcal{S}^*_{u}$}{
            \textit{//$s_{ti}$ resembles another relation $s_{uv}$ between two other objects }\\
            Remove $s_{ti}$ from $\mathcal{S}_t$
        }
        Select $s_{tr}$, the most informative relation, as: 
        	$s_{tr} \leftarrow \argmax_{s_{ti} \in \mathcal{S}_t} c(s_{ti})$ (Eq. \ref{eqn:str})\\
        Generate expression $S$, by forming a simple noun-phrase from the types of $o_{t}$ \& $o_r$, and the category of $s_{tr}$
\end{algorithm}

\subsection{Kunze Et Al.'s Relative Referring Expression Generation (KRREG) Algorithm}
\label{KRREG}

In this section, we briefly describe the algorithm by Kunze et al. \cite{kunze2017spatial} with which we compared our method. This work proposed several methods for generating different types of referring expressions, including five algorithms for generating descriptions (i) involving only the object type (e.g., `the bottle'), (ii) relative referring expressions (e.g., `the bottle to the right of the book'), (iii) set-relative relations (e.g., `the second bottle from the right'), (iv) proximal relations (e.g., `the cup next to the keyboard'), and (v) distal relations (e.g., `the bottle furthest from you'). Because generation of the relative referring expressions (item (ii)) is closest to our work, we implemented that algorithm to compare with our method. The rest of the section describes this algorithm in detail.

In Kunze et al.'s relative referring expression generation algorithm (KRREG), the set of objects $\mathcal{O}$ and their spatial relations in the scene are required as input. For this, as in our method, we employ Mask R-CNN to obtain $\mathcal{O}$ and our RPN network to obtain relations between objects (threshold $T$ in Equation \ref{eqn:rel} is selected as 0.5). 

KRREG's algorithm first defines a set of distractors ($D_t$) as the set of objects with same type of $o_t$:
\begin{equation}
D_t = \{o_i\ |\  n(o_i) = n(o_t)\ \textrm{and}\ o_i \neq o_t , \quad \forall o_i \in \mathcal{O}\},
\end{equation}
and the set of landmarks ($L_t$) as the set of objects in the scene excluding $o_t$ and the ones in $D_t$: 
\begin{equation}
L_t = \{o_i\ |\  o_i \notin D_t \ \textrm{and}\ o_i \neq o_t, \quad \forall o_i \in \mathcal{O}\}.
\end{equation}

In KRREG's algorithm each landmark $o_l \in L_t$ is assigned a rank ($q_l$) measuring its suitability to determine the reference object. The rank of a landmark, $q_l$, is defined to be proportional to $o_l$'s area in the image and inversely proportional to $o_l$'s distance to $o_t$ and the number of objects in $D_l$:
\begin{equation}
    q_l = \dfrac{w^l \times h^l }{d(o_t, o_l)\times |D_l|},
\end{equation}
where $w^l$ and $h^l$ represent the normalized width and height of landmark $o_l$; $|D_l|$ is the number of objects in $D_l$, and $d(o_t, o_l)$ is the distance between $o_t$ and $o_l$ ($<x_c^i, y_c^i>$ denotes the center of mass for object $o_i$):
\begin{equation}
    d(o_t, o_l) = \left( \left(x_c^l - x_c^t\right)^2 + \left(y_c^l - y_c^t\right)^2 \right)^{1/2}.
\end{equation}

After calculating $q_l$ for each landmark $o_l$, $L_t$ is sorted in decreasing order with respect to the $q$ values, and the distinctiveness of each $s_{tl}$ for object $o_l \in L_t$ (or if more than one relation between $o_t$ and $o_l$, each set in the power set of relations) is checked: A candidate relation $s_{tl}$ is regarded as not distinctive if there is a spatial relation $s_{ij}$ for objects $o_i \in D_t$ and $o_j \in L_t$ such that $n(o_l) =  n(o_j)$ and $s_{tl} = s_{ij}$.

While forming a referring expression, KRREG's algorithm assumes an ordering between relations (e.g., if $o_t$ is both `behind' and `to the left' of  $o_l$, `behind' is prior than `to the left') and checks the distinctiveness of relations according to their priority. In our implementation of KRREG, we do the same except for two relations (`close' and `distant') that we do not have in our problem. We replace these two relations with `on top' and `at the bottom'.


\section{Experiments and Results}


We assess our learning-based method in terms of relation estimation accuracies and the \textit{naturalness} of the generated expressions by conducting an evaluation with humans.

\subsection{Training Data}
\label{data}

We use the Visual Genome (VG) dataset \cite{krishnavisualgenome} for training the networks. This dataset contains 108,077 images of indoor as well as outdoor scenes, and 40,480 unique relations (these include other types of relations like affordances, \textit{is-a} information in addition to spatial relations). 


For our first model, i.e., RPN, we collected an arbitrary set of $\{\textrm{input:}\  (o_i, o_j), \textrm{output:}\ r\}$ instances, where $o_i$, $o_j$ and $r$ are as introduced in Section \ref{sect:REG}. In total, we had 5,940 such instances (990 for each relation $r \in \mathcal{R}$) from the VG dataset.

\begin{figure}
\centerline{
\subfigure[Object coordinates.]{
	\includegraphics[width=0.35\textwidth]{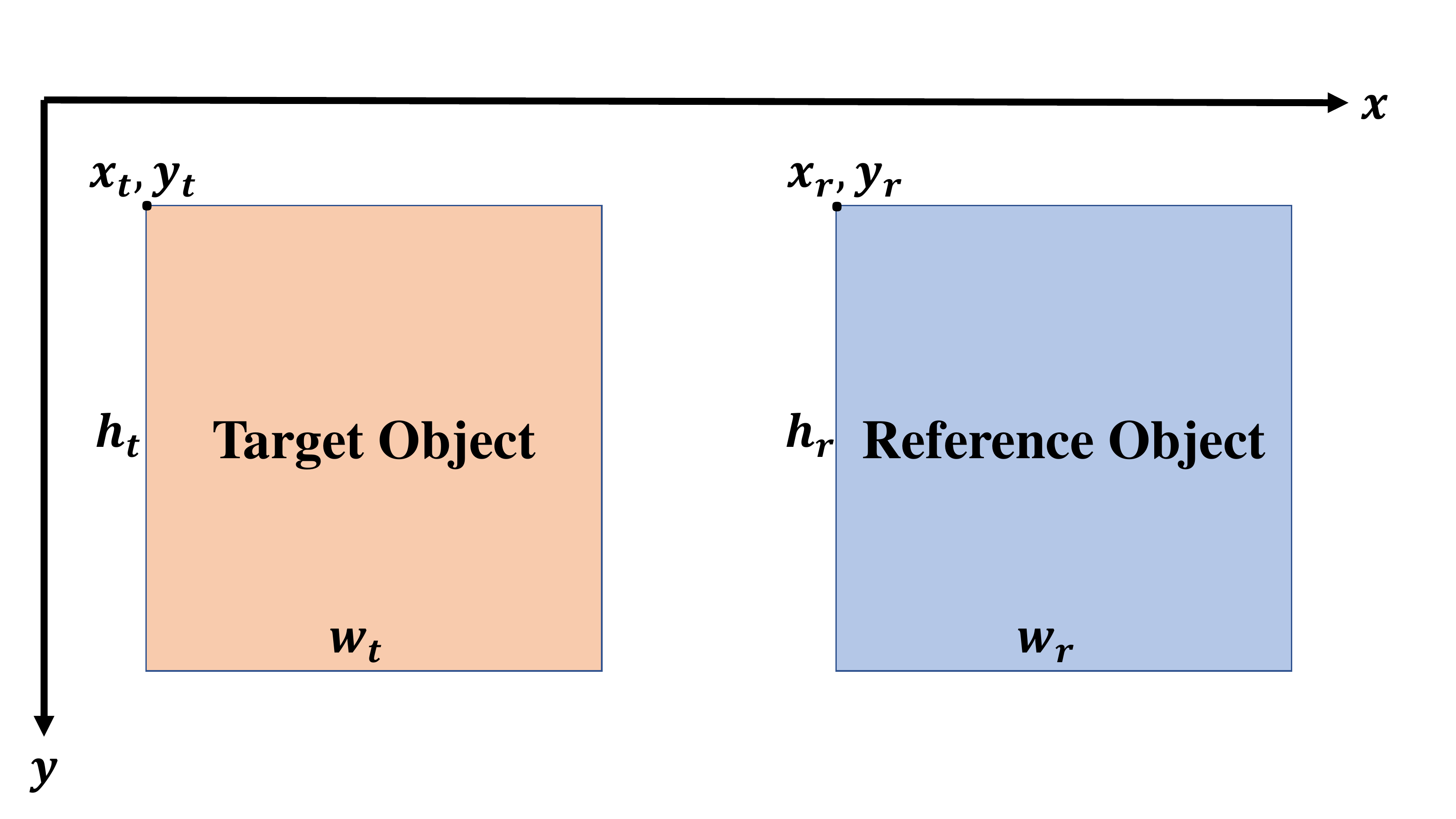}}
}
\centerline{
\footnotesize
\subfigure[Rules for each relation.]{
   \begin{tabular}{|c|c|}\hline \textbf{Relation} & \textbf{Rule for the relation} \\ \hline \hline 
   `to the right' & $x_t > x_r \  and \ (x_t + w_t) > (x_r + w_r) $ \\ \hline
   `to the left' & $x_t < x_r \  and \ (x_t + w_t) < (x_r + w_r) $ \\ \hline 
   `on top' & $x_t > x_r \  and \ (x_t + w_t) < (x_r + w_r) $ \\ \hline
    `at the bottom' & $x_t < x_r \  and \ (x_t + w_t) > (x_r + w_r) $ \\ \hline
   `in front' & $y_t > y_r \  and \ (y_t + h_t) > (y_r + h_r) $ \\ \hline 
    `behind' & $y_t < y_r \  and \ (y_t + h_t) < (y_r + h_r) $ \\ \hline \end{tabular}
   }
}
\caption{Simple logical rules for relations.}
\label{rule_rel}
\end{figure}

For the second network, i.e., RIN, we gathered a set of $\{\textrm{input:}\  (o_i, o_j, r),\ \textrm{output:}\ I(r)\}$ instances, where $I(r)$ is a binary label indicating whether $r$ is an informative relation between $o_i$ and $o_j$ or not. From the VG dataset, we formed 2,057 informative and 2,057 uninformative such instances for each $r \in \mathcal{R}$, yielding a dataset of 24,684 different relation pairs in total. The `informative' relations are determined directly by using the annotations from the VG dataset; i.e., if there is a labeled relation $r$ between objects $o_i$ and $o_j$ in the VG dataset, we set $I(r)$ as `informative'. On the other hand, for collecting `uninformative' relations, we defined simple geometrical rules for each $r \in \mathcal{R}$ -- see Figure \ref{rule_rel}. If a relation $r$ between two objects is suggested by these rules but not annotated in the VG dataset, $I(r)$ is set to `uninformative'. Because humans annotated only the most prominent relations of the scene (rather than all possible ones) in the VG dataset, an unannotated relation between two objects suggests that that relation is not natural for describing these objects. 

\subsection{Analyzing the Networks}
In this section, we analyze the training and testing accuracies of our networks as well as the probability and confidence estimations of the relations. 

\subsubsection{Accuracy Results}
\begin{table}[hbt!]
 	\caption{Training, validation and test accuracies obtained from 10-fold cross-validation. \label{table:acc}}
 	\centering\scriptsize
 	\begin{tabular}{| c | c | c | c |}\hline
	& \textbf{Training} & \textbf{Validation} & \textbf{Testing}\\ \hline \hline 
    \textbf{RPN} & 93.24\% (+/- 0.41\%) & 93.30\% (+/- 0.69)\%  & 93.06\% (+/- 0.63\%)\\ \hline
    \textbf{RIN} & 82.76\% (+/- 0.70\%) &80.18\% (+/- 0.61\%) & 80.28\% (+/- 0.34\%)\\ \hline
 	\end{tabular}
\end{table}

We present the training, validation, and testing performances of RPN and RIN in Table \ref{table:acc}. We observe that the networks do not exhibit any over-fitting because there are no significant differences among the training, validation, and test scores.
Moreover, we note that the accuracies of RIN are lower than those of RPN, suggesting that determining the informativeness of a relation compared to its presence is a more challenging task.

\subsubsection{Analyzing RPN and RIN in detail}

\begin{figure}
\centerline{
 \subfigure[Objects and their bounding boxes.]{
 \includegraphics[width=0.15\textwidth]{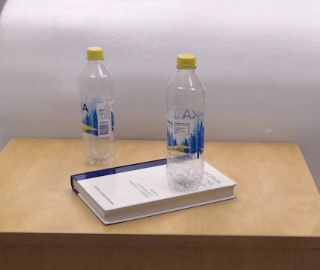}
 \includegraphics[width=0.15\textwidth]{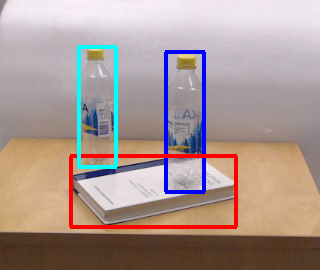}
	\includegraphics[width=0.15\textwidth]{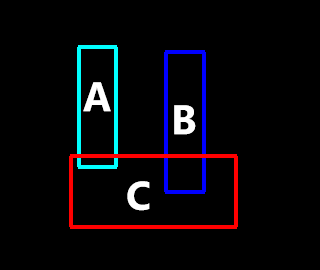}
	} 
}
\centerline{
 \subfigure[Relationship estimation probabilities.]{
\scriptsize
\setlength{\tabcolsep}{4pt}
 	\begin{tabular}{| c | c | c | c | c | c | c  |}\hline
 	    & \multicolumn{6}{|c|}{\textbf{object C}} \\ 
		& \textbf{`right'} & \textbf{`left'} &  \textbf{`on top'} & \textbf{`bottom'} & \textbf{`in front'} &\textbf{`behind'} \\ \hline \hline
   \textbf{object A}   & 23.38\% & 0.33 \% & 0.04\% & 14.87\% & \textbf{61.34\%} & 0.04\%\\ \hline
   \textbf{object B} & 2.71\% &7.10\% & 0.08\% & \textbf{75.89\%} & 13.82\% & 0.40\%\\ \hline
 	\end{tabular}
}
}
\caption{Relations estimated by RPN for referring to the book (object C). RPN chooses different relations for the two bottles although their bounding boxes have very similar placements with respect to the book.}
\label{boxes}
\end{figure}
\begin{figure}[t!]

\centerline{
\subfigure[Objects and their bounding boxes.]{
	\includegraphics[width=0.15\textwidth]{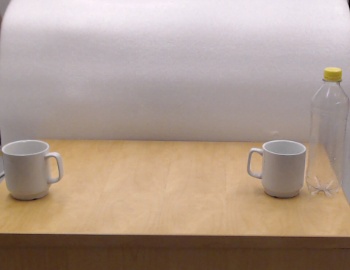}
	\includegraphics[width=0.15\textwidth]{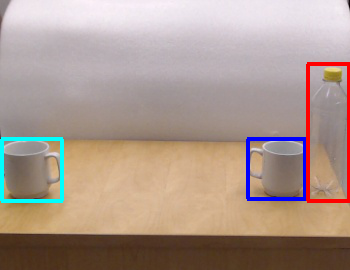}
	\includegraphics[width=0.15\textwidth]{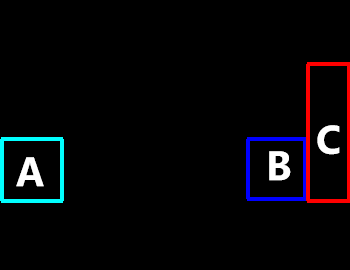}
}
}
\centerline{\scriptsize
\subfigure[Probabilities and confidences of the `to the right' relation.]{
 	\begin{tabular}{| c | c | c |c | }\hline
    & \textbf{Probability ($p(r)$)} & \textbf{Confidence ($c(r)$)} \\ 
    & \textbf{(from RPN)} & \textbf{(from RIN)} \\  \hline\hline
   object C, `to the right', object A   & \textbf{96.49}\% & 49.40\%\\ \hline
   object C, `to the right', object B & 87.35\% & \textbf{98.60}\% \\ \hline
 	\end{tabular}}
}
\caption{Results of RPN and RIN for describing the bottle (object C) with the `to the right' relation (i.e., `the bottle is to the right of object X').}
\label{boxes3}
\end{figure}

In this section, we analyze the performance of RPN and RIN on two example scenes in order to better illustrate the necessity of the two-stage approach and the challenge of the problem.

First, we examine RPN on a challenging example in Figure \ref{boxes} where we want to analyze the relations between the book (object C) and the bottles (object A and object B). In the figure, object C is `in front' of object A, and `at the bottom' of the object B. When we want to describe object C, both of these relations (`in front' and `at the bottom') might be valid for object A and object B if we only consider the bounding boxes of the objects. However, for referring to object C, RPN successfully suggests the `in front' relation for object A and `at the bottom' relation for object B.

In another example in Figure \ref{boxes3}, we analyze and compare the behavior of RPN and RIN. We expect RPN to estimate the relations and RIN to assess their informativeness. To see this visually, `to the right' relation is analyzed to describe the bottle (object C) in Figure \ref{boxes3}. Object A is spatially further away from object C compared to object B, and hence, RPN outputs a higher `to the right' relation probability. However, when the informativeness of this relation is examined, object B has a higher relation confidence (i.e., more informative) because it is closer to object C.


\subsection{User Study}

\begin{figure}
    \centering
    \footnotesize
    \stackunder[5pt]{
    \includegraphics[width=0.25\textwidth]{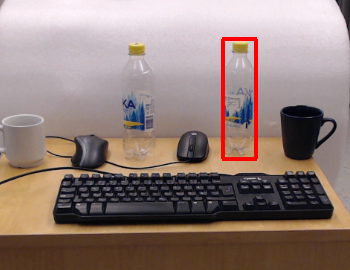}}{\textbf{Our Method:} `The bottle to the left of the cup'}
    \caption{An example scene where the spatial relations between the bottles and other objects  are similar so the KRREG algorithm cannot generate any expression to describe the target.}
    \label{only}
\end{figure}
In this section, we compare our method and KRREG in scenes where referring expressions can be ambiguous (i.e., more than one object may satisfy the description).  For this, we conduct a user study with 61 human judges (27 females and 34 males between 22-41 years old -- average age is $\sim$28) blind to our research questions evaluated the generated expressions from both methods in terms of (i) correctly finding the referenced objects (Experiment 1), and (ii) the naturalness of the expressions (Experiment 2).

For this purpose, we collected a dataset that consisted of three parts: (i) 32 challenging table scenes that we collected from various configurations of objects. The scenes illustrate commonly encountered settings for collaborative tasks performed at a table in order to appropriately test whether a robot in these scenarios would be able to direct people to find the correct object when there are ambiguities. (ii) 20 indoor scenes from 10 different spatial contexts (including a living room, bedroom, furniture store, classroom, playroom, study room, office, bathroom, kitchen, and dining room) from the SUN-RGBD dataset \cite{song2015sun} that contain ambiguities. (iii) 13 ambiguous outdoor scenes from the SUN dataset \cite{xiao2010sun}. Note that none of the models is trained on this dataset. 

Among these 65 scenes, our method and  KRREG generate the same expressions for 35\% of cases (23 scenes), and for 17\% of cases (11 scenes),  KRREG could not generate any expression to describe the target. This happens when none of the relations between landmarks and the target is considered to be distinctive by the  algorithm (distinctiveness  is explained in Section \ref{KRREG}) (see Figure \ref{only} for an example). However, because our model does not consider all spatial relations in the scene as informative, it successfully eliminates the ambiguous ones and describes the target for these 17\% of cases as well. The results reported below show the cases when two methods generate different expressions for the same scene (31 scenes).

\subsubsection{Experiment 1: Correctly finding the referred object}

In this experiment, we compare the methods on their ability to yield unambiguous referring expressions in ambiguous cases. For this purpose, the 61 human judges were asked to select the referred object from the generated referring expressions. They were requested not to select an object if the generated expression is unclear or they were not able to decide on an object according to the generated expression. For a fair comparison, a test instance is provided to a user two times: once for our method and once for the KRREG algorithm. The order of the images with an expression from our method or KRREG was selected randomly by ensuring the same image does not appear successively for different expressions.  



\begin{figure}
    \centering
    \includegraphics[width=0.45\textwidth]{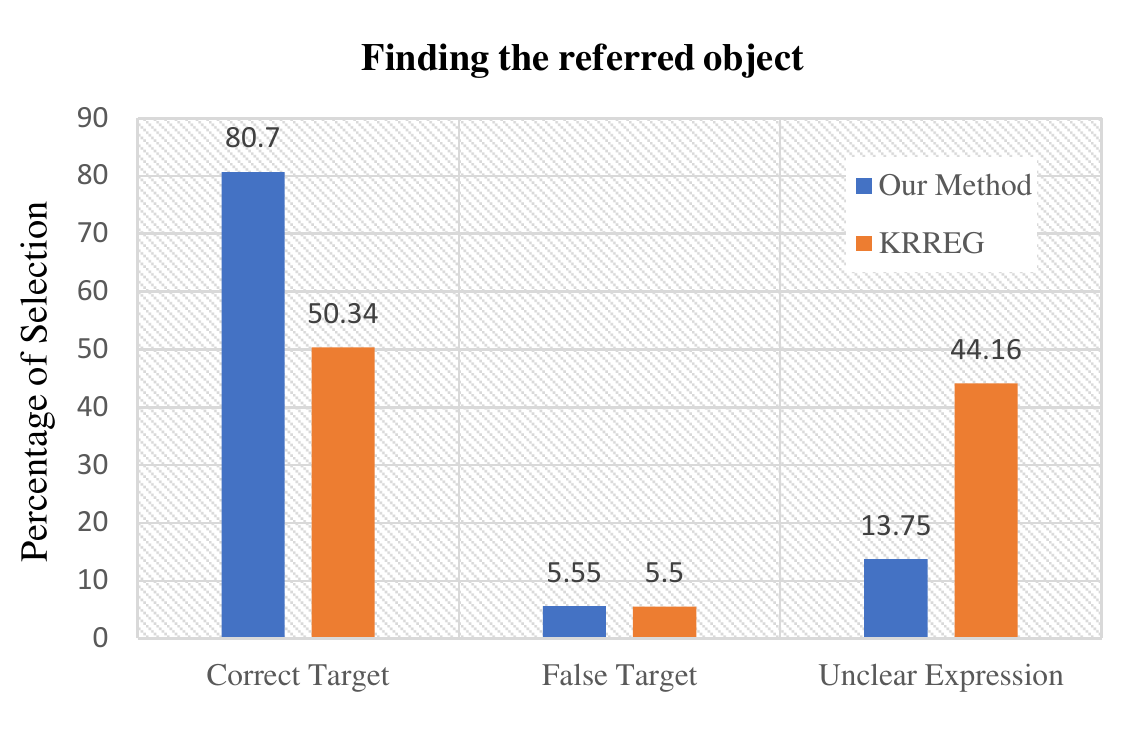}    
    \caption{Qualitative results from Experiment 1: The percentage of cases where human judges correctly selected the referred object. 
    }
    \label{probs_exp1}
\end{figure}

 
The results of this experiment are shown in Figure \ref{probs_exp1}. We observe that the judges were more likely to pick the correct object when the referring expression was generated by our method (80.70\%) compared to KRREG (50.34\%). Moreover, the judges found our generated referring expressions less unclear (13.75\%) than those from the KRREG algorithm (44.16\%). 

To observe whether the method affects the accuracy of the expressions, we analyzed the statistical significance of the evaluations with the Chi-Square test. To do so, we considered the total number of correct, false and unclear selections 
made by judges for the two different methods. This analysis shows that our results are statistically significant,  $\chi^2\ (2,\ N = 3782) = 434.91,\ p < .00001 $.
Therefore, our method is significantly more likely to yield an unambiguous referring expression than the KRREG algorithm.

\subsubsection{Experiment 2: Naturalness of the expressions}


In the first task, we were looking for accuracy of selecting the referred object; in the second task, we wanted to understand whether the expressions generated by our method are considered more natural (i.e., whether people would be more likely to use that expression to refer to the target object) than the KRREG algorithm.
\begin{figure}
\tiny
    \centering
    \includegraphics[width=0.28\textwidth]{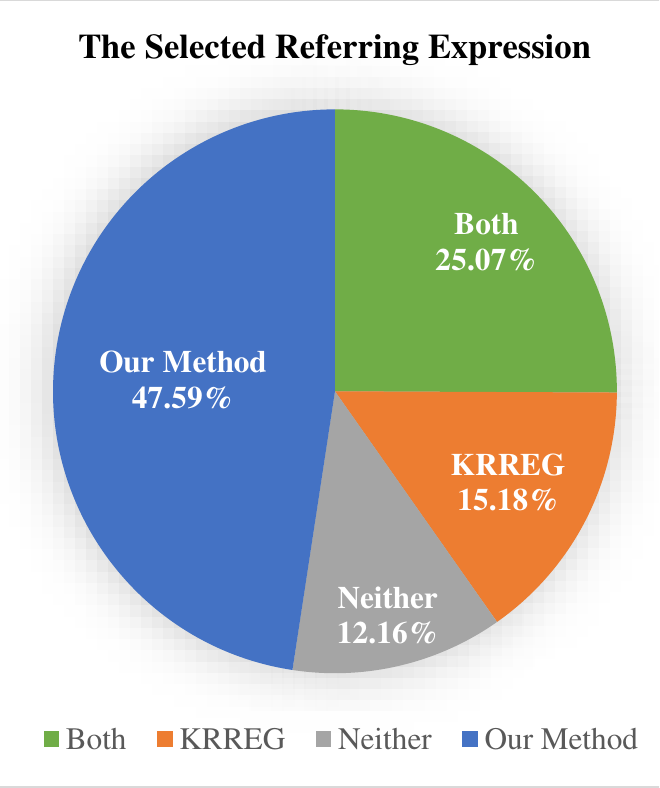} 
    \caption{Qualitative results from Experiment 2: Naturalness of the generated expressions. The percentage of cases where human judges preferred a specific expression for referring to an object. 
    }
    \label{probs_exp2}
\end{figure}
For this purpose, the 61 judges were asked to select which of the two generated referring expressions (one from our method and one from the KRREG algorithm) they would be more likely to use while describing a target object. 
That selection was then considered the more natural of the two. The human judges were also given the option of selecting neither of them or both of them. The order of the expressions generated by two different methods was chosen randomly for each image to avoid bias. 

\begin{figure*}
\centerline{
\subfigure[\textbf{Our:} `The bottle to the right of the mouse' \newline \textbf{KRREG: }`The bottle to the left of the book']{
	\includegraphics[width=0.28\textwidth]{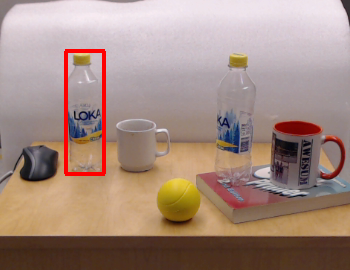}
	}\hspace{1em}
     \subfigure[\textbf{Our:} `The bowl to the left of the vase' \newline \textbf{KRREG: }`The bowl to the right of the oven']{
	\includegraphics[width=0.28\textwidth]{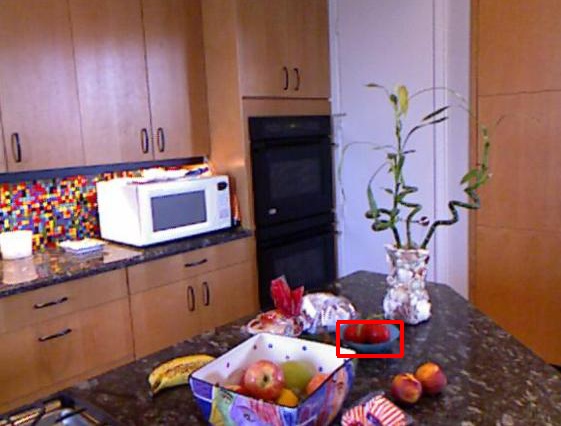}
	} 
\hspace{1em}
   \subfigure[\textbf{Our:} `The car to the right of the person' \newline \textbf{KRREG: }`The car to the left of the vase']{
	\includegraphics[width=0.30\textwidth]{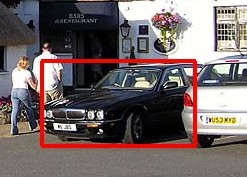}
	}
}
\caption{Examples of the referring expressions generated by our learning-based approach and by the KRREG algorithm for the user study.}
\label{fig:int2}
\end{figure*}

The results of the experiment are presented in Figure \ref{probs_exp2}. We observe that the expressions generated by our method are preferred significantly by the human judges: they selected 47.59\% of our referring expressions and only 15.18\% of the KRREG algorithm expressions. In 25.07\% of the cases, results of both methods were usable, whereas both methods failed to produce suitable results 12.16\% of the time. 

\section{Discussion and Conclusion}

In this paper, we have proposed a learning-based method for estimating informative spatial relations to generate unambiguous and natural referring expressions for collaborative human-robot interaction. Our approach contains two distinct models, one for learning relations between objects and another one to decide a relation's informativeness. 



Through our user study, we have shown that our method is capable of generating unambiguous referring expressions for indoor and outdoor scenes that humans prefer over the state-of-the-art KRREG algorithm. The relation probabilities shown in Figure~\ref{boxes} demonstrate that the proposed RPN is capable of estimating the most dominant relation when only the objects' bounding boxes are provided. Moreover, RIN determines the informativeness of the relations with high accuracy, as shown in Figure~\ref{boxes3}. We have demonstrated in Figure~\ref{only} that using the uniqueness of informative relations can reduce ambiguity in very challenging scenes that the KRREG algorithm fails to generate an expression. Therefore, by utilizing RIN's informativeness measure to select among relations proposed by the RPN, our method generates referring expressions for describing target objects more precisely, as presented in Figure~\ref{probs_exp1}. Moreover, generating the referring expressions with regard to their informativeness yields more natural results, as demonstrated in an evaluation with human judges in Figure~\ref{probs_exp2}.

In summary, we have demonstrated that our method is applicable to different indoor and outdoor environments, which is crucial for use by a robot operating in the real world. Moreover, we have evaluated our results regarding two fundamental aspects of HRI tasks,  communicating unambiguously and naturally in different environments. The promising results indicate that our method can be employed successfully in various collaborative HRI tasks, e.g., helping people to find  ingredients while preparing a recipe or to determine which pieces to use during furniture assembly. 


Our work can be extended in multiple ways. First, more relations, such as `inside' and `next to', can be added to those we considered in this work if we provide additional labeled training data.
Another possibility is to  provide the objects' depth distances, which could improve the training and test accuracies in Table \ref{table:acc}. 
Another avenue worth exploring would be employing RPN to replace the rules in Figure \ref{rule_rel}. Using RPN while labeling the non-informative relations could potentially increase the training and test performances of RIN, and it is part of our future research plans.






\section*{Acknowledgment}
This work was partially funded by a grant from the Swedish Research Council (reg. number 2017-05189). We are grateful to participants for their voluntary contributions to evaluate the referring expressions and Liz Carter for her valuable comments. 

\bibliographystyle{IEEEtran}
\bibliography{references}

\begin{thebibliography}{10}
\providecommand{\url}[1]{#1}
\csname url@rmstyle\endcsname
\providecommand{\newblock}{\relax}
\providecommand{\bibinfo}[2]{#2}
\providecommand\BIBentrySTDinterwordspacing{\spaceskip=0pt\relax}
\providecommand\BIBentryALTinterwordstretchfactor{4}
\providecommand\BIBentryALTinterwordspacing{\spaceskip=\fontdimen2\font plus
\BIBentryALTinterwordstretchfactor\fontdimen3\font minus
  \fontdimen4\font\relax}
\providecommand\BIBforeignlanguage[2]{{%
\expandafter\ifx\csname l@#1\endcsname\relax
\typeout{** WARNING: IEEEtran.bst: No hyphenation pattern has been}%
\typeout{** loaded for the language `#1'. Using the pattern for}%
\typeout{** the default language instead.}%
\else
\language=\csname l@#1\endcsname
\fi
#2}}

\bibitem{hurford2007semantics}
J.~R. Hurford, B.~Heasley, and M.~B. Smith, \emph{Semantics: a
  coursebook}.\hskip 1em plus 0.5em minus 0.4em\relax Cambridge University
  Press, 2007.

\bibitem{hatori2018interactively}
J.~Hatori, Y.~Kikuchi, S.~Kobayashi, K.~Takahashi, Y.~Tsuboi, Y.~Unno, W.~Ko,
  and J.~Tan, ``Interactively picking real-world objects with unconstrained
  spoken language instructions,'' in \emph{ICRA}.\hskip 1em plus 0.5em minus
  0.4em\relax IEEE, 2018.

\bibitem{shridhar2017grounding}
M.~Shridhar and D.~Hsu, ``Grounding spatio-semantic referring expressions for
  human-robot interaction,'' \emph{RSS Workshop on Spatial-Semantic
  Representations in Robotics}, 2017.

\bibitem{shridhar2018interactive}
------, ``Interactive visual grounding of referring expressions for human-robot
  interaction,'' in \emph{RSS}, 2018.

\bibitem{williams2017referring}
T.~Williams and M.~Scheutz, ``Referring expression generation under
  uncertainty: Algorithm and evaluation framework,'' in \emph{INLG}, 2017.

\bibitem{williamsreferring}
------, ``Referring expression generation under uncertainty in integrated robot
  architectures,'' in \emph{RSS Workshop on Human-Centered Robotics:
  Interaction, Physiological Integration and Autonomy}, 2017.

\bibitem{kunze2017spatial}
L.~Kunze, T.~Williams, N.~Hawes, and M.~Scheutz, ``Spatial referring expression
  generation for hri: Algorithms and evaluation framework,'' in \emph{AAAI Fall
  Symposium on AI and HRI}, 2017.

\bibitem{zender2009situated}
H.~Zender, G.-J.~M. Kruijff, and I.~Kruijff-Korbayov{\'a}, ``Situated
  resolution and generation of spatial referring expressions for robotic
  assistants,'' in \emph{IJCAI}, 2009.

\bibitem{mao2016generation}
J.~Mao, J.~Huang, A.~Toshev, O.~Camburu, A.~L. Yuille, and K.~Murphy,
  ``Generation and comprehension of unambiguous object descriptions,'' in
  \emph{CVPR}.\hskip 1em plus 0.5em minus 0.4em\relax IEEE, 2016.

\bibitem{yu2017joint}
L.~Yu, H.~Tan, M.~Bansal, and T.~L. Berg, ``A joint speakerlistener-reinforcer
  model for referring expressions,'' in \emph{CVPR}.\hskip 1em plus 0.5em minus
  0.4em\relax IEEE, 2017.

\bibitem{cirik2018using}
V.~Cirik, T.~Berg-Kirkpatrick, and L.-P. Morency, ``Using syntax to ground
  referring expressions in natural images,'' \emph{AAAI}, 2018.

\bibitem{kruijff2007incremental}
G.-J.~M. Kruijff, P.~Lison, T.~Benjamin, H.~Jacobsson, and N.~Hawes,
  ``Incremental, multi-level processing for comprehending situated dialogue in
  human-robot interaction,'' in \emph{Symposium on Language and Robots}, 2007.

\bibitem{kollar2013learning}
T.~Kollar, V.~Perera, D.~Nardi, and M.~Veloso, ``Learning environmental
  knowledge from task-based human-robot dialog,'' in \emph{ICRA}.\hskip 1em
  plus 0.5em minus 0.4em\relax IEEE, 2013.

\bibitem{paul2016efficient}
R.~Paul, J.~Arkin, N.~Roy, and T.~M~Howard, ``Efficient grounding of abstract
  spatial concepts for natural language interaction with robot manipulators,''
  in \emph{RSS}, 2016.

\bibitem{dale1995computational}
R.~Dale and E.~Reiter, ``Computational interpretations of the gricean maxims in
  the generation of referring expressions,'' \emph{Cognitive science}, vol.~19,
  no.~2, pp. 233--263, 1995.

\bibitem{viethen2008use}
J.~Viethen and R.~Dale, ``The use of spatial relations in referring expression
  generation,'' in \emph{INLG}.\hskip 1em plus 0.5em minus 0.4em\relax
  Association for Computational Linguistics, 2008.

\bibitem{moratz2006spatial}
R.~Moratz and T.~Tenbrink, ``Spatial reference in linguistic human-robot
  interaction: Iterative, empirically supported development of a model of
  projective relations,'' \emph{Spatial cognition and computation}, vol.~6,
  no.~1, pp. 63--107, 2006.

\bibitem{guadarrama2013grounding}
S.~Guadarrama, L.~Riano, D.~Golland, D.~Go, Y.~Jia, D.~Klein, P.~Abbeel,
  T.~Darrell, \emph{et~al.}, ``Grounding spatial relations for human-robot
  interaction,'' in \emph{IROS}.\hskip 1em plus 0.5em minus 0.4em\relax IEEE,
  2013.

\bibitem{skubic2004spatial}
M.~Skubic, D.~Perzanowski, S.~Blisard, A.~Schultz, W.~Adams, M.~Bugajska, and
  D.~Brock, ``Spatial language for human-robot dialogs,'' \emph{IEEE SMC, Part
  C (Applications and Reviews)}, vol.~34, no.~2, pp. 154--167, 2004.

\bibitem{zhang2009rule}
C.~Zhang, X.~Zhang, W.~Jiang, Q.~Shen, and S.~Zhang, ``Rule-based extraction of
  spatial relations in natural language text,'' in \emph{CiSE}.\hskip 1em plus
  0.5em minus 0.4em\relax IEEE, 2009.

\bibitem{viethen2013graphs}
J.~Viethen, M.~Mitchell, and E.~Krahmer, ``Graphs and spatial relations in the
  generation of referring expressions,'' in \emph{Proceedings of the 14th
  European Workshop on Natural Lang. Gen.}, 2013, pp. 72--81.

\bibitem{malinowski2014pooling}
M.~Malinowski and M.~Fritz, ``A pooling approach to modelling spatial relations
  for image retrieval and annotation,'' \emph{arXiv:1411.5190}, 2014.

\bibitem{haldekar2017identifying}
M.~Haldekar, A.~Ganesan, and T.~Oates, ``Identifying spatial relations in
  images using convolutional neural networks,'' in \emph{IJCNN}.\hskip 1em plus
  0.5em minus 0.4em\relax IEEE, 2017.

\bibitem{huang2017speed}
J.~Huang, V.~Rathod, C.~Sun, M.~Zhu, A.~Korattikara, A.~Fathi, I.~Fischer,
  Z.~Wojna, Y.~Song, S.~Guadarrama, \emph{et~al.}, ``Speed/accuracy trade-offs
  for modern convolutional object detectors,'' in \emph{CVPR}.\hskip 1em plus
  0.5em minus 0.4em\relax IEEE, 2017.

\bibitem{nair2010rectified}
V.~Nair and G.~E. Hinton, ``Rectified linear units improve restricted boltzmann
  machines,'' in \emph{ICML}, 2010.

\bibitem{srivastava2014dropout}
N.~Srivastava, G.~Hinton, A.~Krizhevsky, I.~Sutskever, and R.~Salakhutdinov,
  ``Dropout: a simple way to prevent neural networks from overfitting,''
  \emph{JMLR}, vol.~15, no.~1, pp. 1929--1958, 2014.

\bibitem{krishnavisualgenome}
R.~Krishna, Y.~Zhu, O.~Groth, J.~Johnson, K.~Hata, J.~Kravitz, S.~Chen,
  Y.~Kalantidis, L.-J. Li, D.~A. Shamma, M.~Bernstein, and L.~Fei-Fei, ``Visual
  genome: Connecting language and vision using crowdsourced dense image
  annotations,'' \emph{IJCV}, vol. 123, no.~1, pp. 32--73, 2017.

\bibitem{song2015sun}
S.~Song, S.~P. Lichtenberg, and J.~Xiao, ``Sun rgb-d: A rgb-d scene
  understanding benchmark suite,'' in \emph{CVPR}.\hskip 1em plus 0.5em minus
  0.4em\relax IEEE, 2015.

\bibitem{xiao2010sun}
J.~Xiao, J.~Hays, K.~A. Ehinger, A.~Oliva, and A.~Torralba, ``Sun database:
  Large-scale scene recognition from abbey to zoo,'' in \emph{CVPR}.\hskip 1em
  plus 0.5em minus 0.4em\relax IEEE, 2010.

\end{thebibliography}

\end{document}